\UseRawInputEncoding
\pdfoutput=1
\documentclass[final,5p,times,twocolumn,authoryear]{elsarticle}
\usepackage{amsmath}
\usepackage{amsfonts}
\usepackage{algorithm}
\usepackage{array}
\usepackage[caption=false,font=normalsize,labelfont=sf,textfont=sf]{subfig}
\usepackage{textcomp}
\usepackage{stfloats}
\usepackage{url}
\usepackage[colorlinks,linkcolor=red]{hyperref}
\usepackage{verbatim}
\usepackage{graphicx}
\usepackage{booktabs}
\usepackage{color}
\usepackage{bbm, dsfont}
\usepackage[noend]{algpseudocode}
\journal{Expert Systems with Applications}

\begin{document}
\begin{frontmatter}

\title{FireMatch: A Semi-Supervised Video Fire Detection Network Based on Consistency  and  Distribution Alignment}

\author[1,2]{Qinghua Lin}
\ead{akametris@163.com}
\author[2]{Zuoyong Li\corref{corr}}
\ead{fzulzytdq@126.com}
\cortext[corr]{Corresponding author.}
\author[2]{Kun Zeng}
\ead{zengkun@mju.edu.cn}
\author[3]{Haoyi Fan}
\ead{fanhaoyi@zzu.edu.cn}
\author[1]{Wei Li}
\ead{liwei33660@163.com}
\author[4,5]{Xiaoguang Zhou}
\ead{zxg@bupt.edu.cn}

\address[1]{School of Computer Science and Mathematics, Fujian University of Technology, 350118, Fuzhou, China \\}
\address[2]{Fujian Provincial Key Laboratory of Information Processing and Intelligent Control, College of
Computer and Control Engineering, Minjiang University, 350121, Fuzhou, China \\}
\address[3]{School of Computer and Artificial Intelligence, Zhengzhou University, 450001, Zhengzhou, China \\}
\address[4]{School of Automation, Beijing University of Post and Telecommunications, 100876, Beijing, China}
\address[5]{School of Economics and Management, Minjiang University, 350121, Fuzhou, China}

\begin{abstract}
Deep learning techniques have greatly enhanced the performance of fire detection in videos. However, video-based fire detection models heavily rely on labeled data, and the process of data labeling is particularly costly and time-consuming, especially when dealing with videos.
Considering the limited quantity of labeled video data, we propose a semi-supervised fire detection model called FireMatch, which is based on consistency regularization and adversarial distribution alignment. 
Specifically, we first combine consistency regularization with pseudo-label.
For unlabeled data, we design video data augmentation to obtain corresponding weakly augmented and strongly augmented samples. The proposed model predicts weakly augmented samples and retains pseudo-label above a threshold, while training on strongly augmented samples to predict these pseudo-labels for learning more robust feature representations.
Secondly, we generate video cross-set augmented samples by adversarial distribution alignment to expand the training data and alleviate the decline in classification performance caused by insufficient labeled data.
Finally,  we introduce a fairness loss to help the model produce diverse predictions for input samples, thereby addressing the issue of high confidence with the non-fire class in fire classification scenarios.
The FireMatch achieved an accuracy of 76.92\% and 91.81\% on two real-world fire datasets, respectively. The experimental results demonstrate that the proposed method outperforms the current state-of-the-art semi-supervised classification methods. 
\end{abstract}


\end{frontmatter}


\section{Introduction}
Fire is one of the most dangerous disasters due to its rapid spread and destructive power \cite{ZHF23}. It causes not only casualties, but also property damage, environmental pollution, and social impacts. Every year, numerous fire incidents occur globally, many of which are caused by human or natural factors. According to a research report in 2021 \cite{JM21}, there were more than 77,000 forest fire incidents in the central region of India from 2001 to 2022. The repeated occurrences of these fires pose a great health threat to nearby residents with respiratory diseases and have serious impacts on the local ecology and economic development. In addition, the occurrence frequency of building fires is much higher than that of forest fires. According to the International Association of Fire and Rescue Services, there were 23,535 building fire incidents recorded in 18 cities globally in 2017, and from 2013 to 2017, there were 6,581 fire-related casualties in 44 cities worldwide, with electrical fires being one of the main causes of building fires worldwide \cite{GA20}. In 2020, a serious fire accident occurred in a warehouse facility storing ammonium nitrate at the Beirut port in Lebanon due to fireworks, and the subsequent explosion caused 178 deaths, over 6,500 injuries, and more than 300,000 people homeless \cite{SS21}. In most cases, timely detection of indoor fires can greatly reduce casualties and property damage, and early fire detection methods in various forms are possible.

In the past, researchers have explored different approaches for fire detection, utilizing various types of sensors including temperature sensors, smoke sensors, and particle sensors. These sensors have found widespread applications across different scenarios. However, these physical detection sensors are limited by cost and cannot be deployed on a large scale in indoor and outdoor scenes (e.g., physical sensors are difficult to deploy on a large scale in forests or super-large factories). In addition, these fire detection schemes require proximity to the location of the fire to achieve relatively effective early warning and require manual intervention to ensure the authenticity of the warning. Recently, more and more researchers are applying computer vision technology to fire detection scenes \cite{SJ17,LS20,PL20}. Compared with traditional physical sensors, vision-based fire detection schemes have the advantages of wide coverage, low deployment cost, short response time, and strong robustness.

The development of deep learning (DL) has facilitated higher accuracy and widespread application in vision-based fire detection \cite{WZ22,AJS22,JSS23}. However, as a data-driven end-to-end learning technique, DL-based fire detection methods usually require a large number of samples with label for training (i.e. fully supervised learning). But data labeling is time-consuming and labor-intensive, especially for video data. Additionally, due to the imbalance between labeled and unlabeled data, there is often a sampling bias in semi-supervised classification tasks, resulting in a mismatch in empirical distributions and consequently a decline in classification performance. 

To address these challenges, we propose a semi-supervised fire detection method based on consistency regularization with adversarial distribution alignment, called FireMatch. Firstly, the proposed method combines consistency regularization with pseudo-label to predict weakly augmented views of unlabeled video data and retains pseudo-label that exceed a threshold. The model learns robust feature representations by predicting the pseudo-labels of these strongly augmented samples. 
Then, we introduce a self-adaptive threshold that starts low in the initial training phase to obtain as many pseudo-labeled samples as possible and accelerate convergence. As the training progresses, the threshold increases to eliminate erroneous pseudo-labels and improve the model's classification performance. 
Next, to fully utilize both labeled and unlabeled data, we apply adversarial distribution alignment to generate video cross-set augmentation samples, expanding the training samples and simultaneously bridging the gap between the empirical distributions of labeled and unlabeled data.
Finally, inspired by the work \cite{WY22}, we introduce a fairness loss to encourage the model to make diverse predictions on input samples and alleviate the impact of overconfidence in non-fire class caused by data imbalance.

Our contributions can be summarized as follows:

\begin{itemize}
\item To fully leverage unlabeled data, we combine consistency regularization with self-adaptive pseudo-label on video data.  Generating enough pseudo-labeled data with high-quality for training that helps the model in achieving accurate fire video classification.
\end{itemize}

\begin{itemize}
\item For addressing the problem of imbalanced labeled and unlabeled data leading to mismatched sampling experiences, we propose Video Cross-set Sample Augmentation (VCSA) combined with adversarial distribution alignment to generate additional labeled samples and alleviate this bias.
\end{itemize}

\begin{itemize}
\item A fairness loss is introduced to help the model in making more diverse predictions, alleviating the issue of overconfidence in the non-fire class in video fire classification caused by data imbalance.
\end{itemize}

\begin{itemize}
\item We conduct extensive experiments and ablation studies on public datasets and compare our method with state-of-the-art semi-supervised methods. The experimental results demonstrate the effectiveness of the proposed method.
\end{itemize}

\section{Related Work}
Before introducing our method, we first review some previous works related to fire detection, and then give an overview of some semi-supervised classification methods.

\subsection{Fire Detection}
Among various disasters, fire is one of the most frequent and common disasters that threaten public safety and social development. In recent years, researchers have applied deep learning techniques to fire detection scenarios, achieving fire detection models that are more sensitive than physical sensors in flame image segmentation \cite{NM21,HH20,MWS21}, fire detection \cite{HW22,LZ22,AA20}, and fire image classification \cite{HKS22,HC18}. For smoke and fire detection, FireNet \cite{JA19} develop a fire detection unit with Internet of Things (IoT) capabilities, effectively alleviating the problems of false triggering and delayed triggering of physical fire detectors. Barmpoutis et al. \cite{BP19} first use Faster R-CNN \cite{RS15} to detect candidate regions, and then validate the detected fire regions through the analysis of spatial features using a linear dynamical system. Fire HRR \cite{WZ22} achieves real-time prediction of heat release rate by monitoring the behavior of external smoke in building fires, thus identifying the development process of building fires with higher stability. EdgeFireSmoke \cite{AJS22} deploys CNN on edge computing devices for image processing, enabling timely fire alarms with a response time of approximately 30 milliseconds and achieving high accuracy. Li et al. \cite{PL20}  propose a novel fire detection model based on YOLOv3 \cite{RJ18}, which further enhances the fire detection capability of the model. For fire image classification, Sharma et al. \cite{SJ17} explore the performance of CNN on imbalanced fire data and improve the classification accuracy on challenging data using pre-trained VGG16 \cite{SK14} and ResNet-50 \cite{HK16}. DFAN \cite{YH22} focuses on balancing computational cost and accuracy by using spatial attention to capture spatial details for improving fire and non-fire recognition ability, and employing meta-heuristic method to discard redundant parameters. Jandhyala et al. \cite{JSS23} combine Inception-V3 \cite{SC16} with single shot detector for the classification of fire or smoke in aerial images. EFDNet \cite{LS20} extracts multi-scale features to enhance spatial details, and selectively emphasizes the contributions of different feature maps using channel attention mechanism. Jabnouni et al. \cite{JS22} utilize transfer learning on various state-of-the-art deep learning models for fire image classification, where ResNet-50 \cite{HK16} achieves the best performance on a carefully designed dataset. 

The above fire detection solutions rely heavily on labeled image data to improve the model's ability to detect flames. However, video data labeling is costly. Therefore, it is of great significance to study fire classification models based on semi-supervised learning.

\begin{figure*}[!t]
  \centering
  \includegraphics[width=\textwidth]{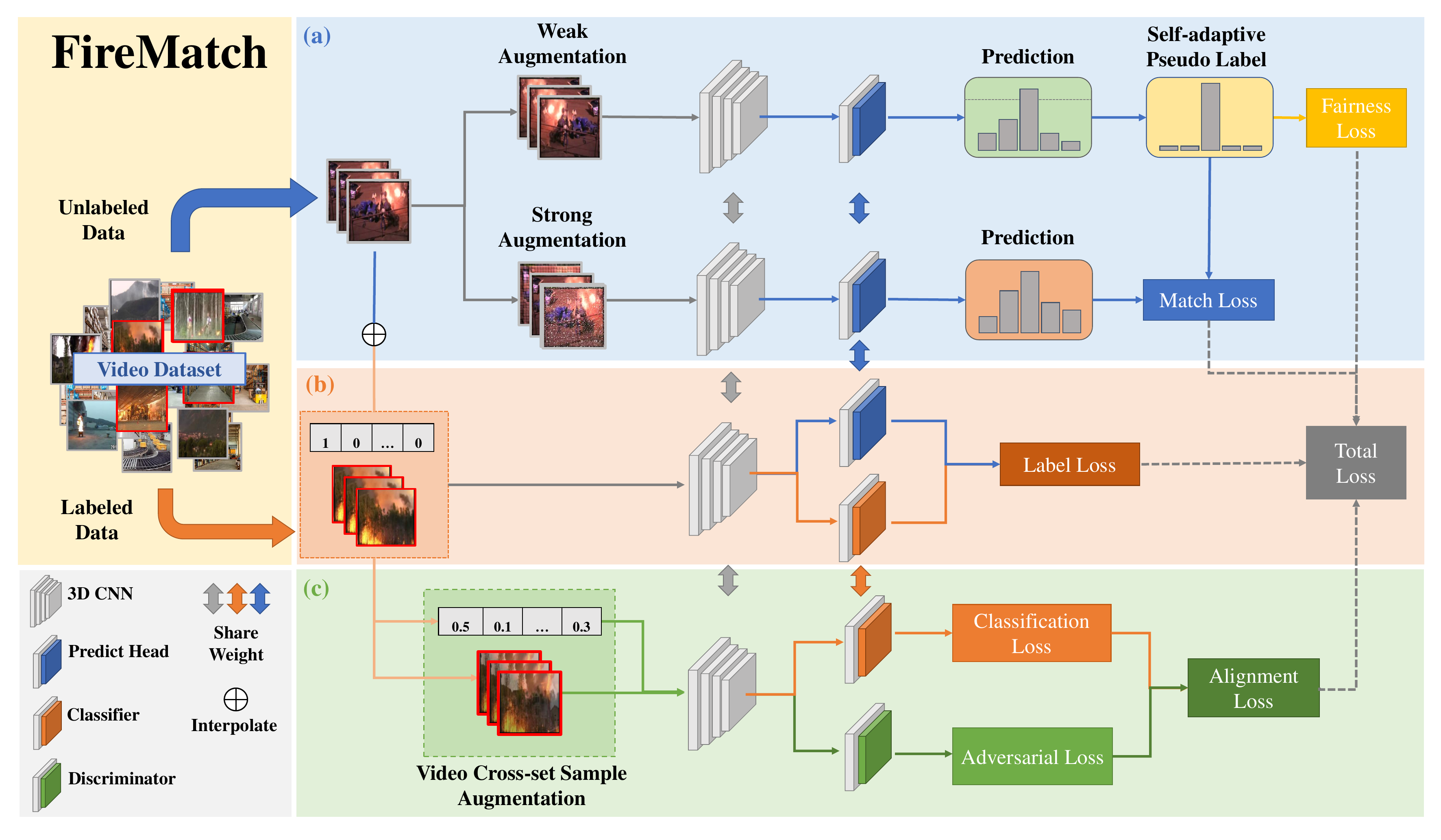}
  \caption{Overview of proposed method FireMatch. (a) Consistency Regularization. Learning more robust feature representations by matching the strong and weak augmented views of unlabeled samples. (b) Supervised Training. The model gains initial classification capability through training on labeled samples. (c) Adversarial Distribution Alignment. By generating video cross-set augmented samples, the generalization capability of the model is further enhanced.}
  \label{Overview}
\end{figure*}

\subsection{Semi-supervised Learning Classification}
In recent years, semi-supervised learning has been widely studied by scholars due to its weak dependence on labeled data compared to supervised learning. Pseudo-labeling (PL) has been maturely applied in semi-supervised learning. The PL-based method obtains an initial model trained on labeled samples to predict the unlabeled samples, and then retrains the model using the predicted results as the labels for the unlabeled samples. For example, Pseudo-Label \cite{LDH13} continues training based on the pseudo-labeled data generated by the previous iteration of the model. Meta Pseudo Labels \cite{PH21} employs a teacher network that constantly adjusts the pseudo-labels based on the student's performance on the labeled data to generate better pseudo-labels. UPS \cite{RMN21} argues that  performence of PL is limited by erroneous high-confidence predictions, and improves the accuracy of pseudo-labels by significantly reducing noise during training through an uncertainty-aware pseudo-label selection method. Zhang et al. \cite{ZX21} suggest using cluster consistency to estimate the similarity of pseudo-labels between consecutive training iterations and refine the pseudo-labels through temporal propagation. As a classic semi-supervised learning technology, consistency regularization assumes that the predicted results should not change significantly when adding some perturbations to unlabeled data. For example, MixMatch \cite{BD19} combines labeled and unlabeled data as augmented data, and guesses low-entropy labels for these unlabeled augmented data. ReMixMatch \cite{BD20} encourages the marginal distribution of unlabeled data to be close to the real distribution through distribution alignment, and matches multiple augmented versions of data to their weakly augmented predictions using an anchor. FixMatch \cite{SK20} encourages the model to generate high-confidence pseudo-labels for weakly augmented data and trains on strongly augmented versions of the same image with these pseudo-labels. FlexMatch \cite{ZB21} proposes the Curriculum Pseudo Labeling (CPL) to flexibly adjust the thresholds of different classes to filter out erroneous pseudo-labeled data. FreeMatch \cite{WY22} introduces an adaptive class fair regularization penalty, which adjusts the confidence threshold of pseudo-labels according to the learning state of the model. SoftMatch \cite{CH23} proves that thresholded pseudo-labeling methods have a quantity-quality trade-off issue, and maintains high-quality pseudo-labels with high quantity by weighting confidence samples during training. SimMatch \cite{ZM22} applies consistency regularization in the semantic level and instance level, encouraging different augmented views of the same instance to have the same class prediction and similar relationship to other instances. SelfMatch \cite{KB21} combines contrastive self-supervised pre-training and consistency regularization for semi-supervised fine-tuning, narrowing the performance gap between supervised learning and semi-supervised learning.

Although current semi-supervised learning tasks can effectively utilize large amounts of unlabeled data to assist model training, they overlook the role of labeled data in generating informative data. Furthermore, there is a scarcity of video classification methods based on semi-supervised learning at present, and the classification performance often falls short of satisfactory results. In this paper, we propose a novel semi-supervised video classification method. Section 3 will provide a detailed introduction to our method.

\section{Method}

\subsection{Network Architecture}
The proposed method is illustrated in Figure \ref{Overview}, when a video dataset $X$ is given, which contains a large amount of unlabeled data ${{\cal U}} = \left\{ {{u_b},b \in (1,...,\mu B)} \right\}$ and a small amount of labeled data ${{\cal X}} = \left\{ {\left( {{x_b},{y_b}} \right),b \in (1,...,B)} \right\}$. $B$ represents the batch size, $\mu$ is a hyperparameter that determines the relative batch sizes of ${{\cal X}}$ and ${{\cal U}}$. ${u_b} \in {\mathbb{R}^{C \times T \times W \times H}}$ and ${x_b} \in {\mathbb{R}^{C \times T \times W \times H}}$ represent the unlabeled and labeled samples, respectively ($C \in {\mathbb{N}^ + }$ represents the number of channels,  $T \in {\mathbb{N}^ + }$ denotes the time and $W, H \in {\mathbb{N}^ + }$ represents the width and height of each frame). ${y_b} \in \left\{ {0,1, \cdots ,{N}} \right\}$ denotes the ground truth labels, where $N$ represents the number of classes.

For each input ${u_b}$, a corresponding pseudo label ${y'_b} \in \left\{ {0,1, \ldots ,N} \right\}$ will be generated based on the results of the earlier iterations of the model. Firstly, for labeled data ${\mathcal{X}}$, the video data features are extracted using a 3D CNN and used to train a classifier ${\mathcal{C}}$ as well as a prediction head ${\mathcal{P}}$. Secondly, for a large amount of unlabeled data ${{\cal U}}$, strong and weak augmentations are applied, and the features are extracted using a 3D CNN with shared weights. Consistency regularization is utilized to match the predictions of the strong augmentations with the adaptive pseudo labels generated from the weak augmentations. Finally, to fully utilize the large amount of unlabeled data and the informative annotated data, we generate interpolated augmented data which are more informative than the unlabeled data by aligning the distributions of the annotated and unlabeled data. The features are extracted using a shared-weight 3D CNN, and the classification is performed by the classifier ${\mathcal{C}}$ while the distribution distance is minimized by the discriminator ${\mathcal{D}}$. This method achieves effective semi-supervised fire video classification. Next, we will cover consistency regularization and adversarial distribution alignment in more detail.

\subsection{Consistency Regularization}
\subsubsection{Strong and Weak Augmentation}

\begin{figure}[!t]
\centering
\includegraphics[width=3.2in]{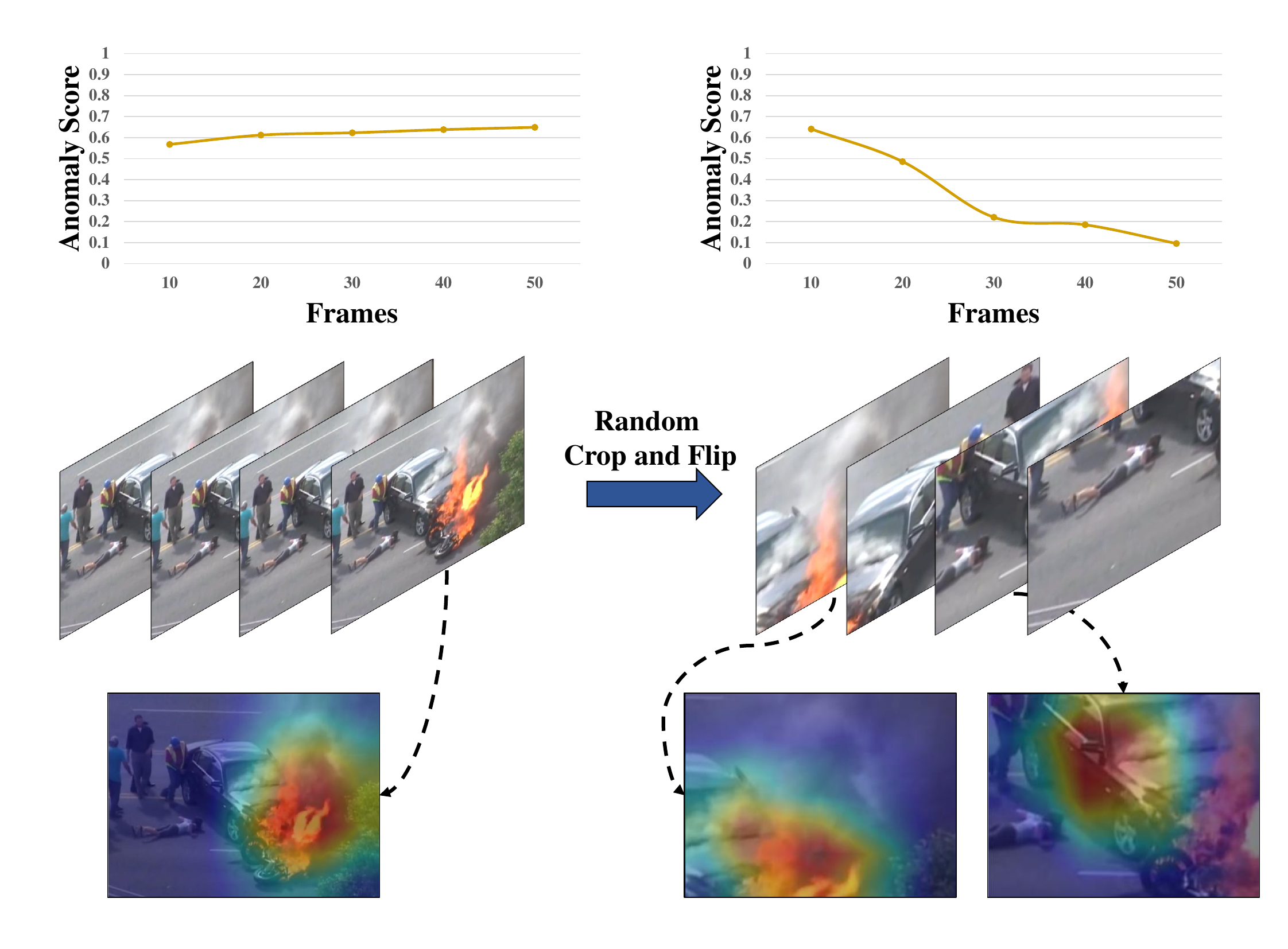}
\caption{(Top Row) The anomaly scores of the video frames. (Middle Row) Original samples and the augmented samples with random crop and flip. (Bottom Row) The class activation mapping (CAM) \cite{ZB16} score maps of the corresponding video frames.}
\label{Random cropping}
\end{figure}

One of the core ideas of consistency regularization is to add some perturbations to the data and encourage the model to produce the same output distribution \cite{BD19}. When adding perturbations to 2D images, common regularization techniques include rotation, flipping, random cropping, sharpening, and so on. However, when dealing with video data, a problem to be addressed is that videos are composed of individual frames, and the arrangement of frames carries important temporal information. Therefore, applying the 2D image augmentation methods crudely to video frames would break the temporal relationships between frames. As shown in Figure \ref{Random cropping}, applying random cropping and flipping to each video frame can cause significant disturbance to the data. So it's not suitable for weak augmentation because the purpose of weak augmentation is to add slight perturbations to the samples without affecting the feature representation. This viewpoint will be demonstrated in the ablation study.

Therefore, we suggest applying only flipping for weak augmentation on video samples and define weak augmentation as $w\left(  \cdot  \right)$. Specifically, for an unlabeled sample ${u_b}$, we define it as ${u_b} = \left\{ {u{x_t}{\text{ , }} t \in \left( {1,...,T} \right)} \right\}$ where $u{x_t}$ represents video frames and $T$ is the total number of frames. Then, we can obtain weak augmented unlabeled samples $w{u_b} = \left\{ {w(u{x_t}{\text{) , }}t \in \left( {1,...,T} \right)} \right\}$. Similarly, we use $\Omega \left(  \cdot  \right)$ to represent strong augmentation. So, the strong augmentation samples can be defined as $s{u_b} = \left\{ {\Omega (u{x_t}{\text{) , }}{}t \in \left( {1,...,T} \right)} \right\}$. The work \cite{SK20} has demonstrated that applying augmentation strategies learned from limited labeled data can be problematic. Therefore, in this work, we adopt RandAugment \cite{CED20} as the strong augmentation strategy for video frames. The RandAugment randomly samples from a predefined range to control the degree of distortion for all samples. Using the aforementioned strong and weak augmentation strategies, we perturb the original data and obtain two sets of augmented samples, denoted as strong augmentation samples $s{u_b}$ and weak augmentation samples $w{u_b}$. These samples are fed into a shared 3D CNN to extract features, and prediction head ${\mathcal{P}}$ is used to output the predicted labels. According to the core idea of feature consistency regularization, prediction head ${\mathcal{P}}$ should output the same class distribution for both strong and weak augmentations. Therefore, we generate pseudo-labels for the weak augmentation samples, and only retained those whose confidence exceeded a certain threshold. Subsequently, the model is trained to match the predictions of the strong augmentation samples and the manually generated pseudo-labels, thereby achieving feature learning on the unlabeled data.

\subsubsection{Self-adaptive Pseudo Label}
\begin{figure*}
  \centering
  \includegraphics[width=0.9\textwidth]{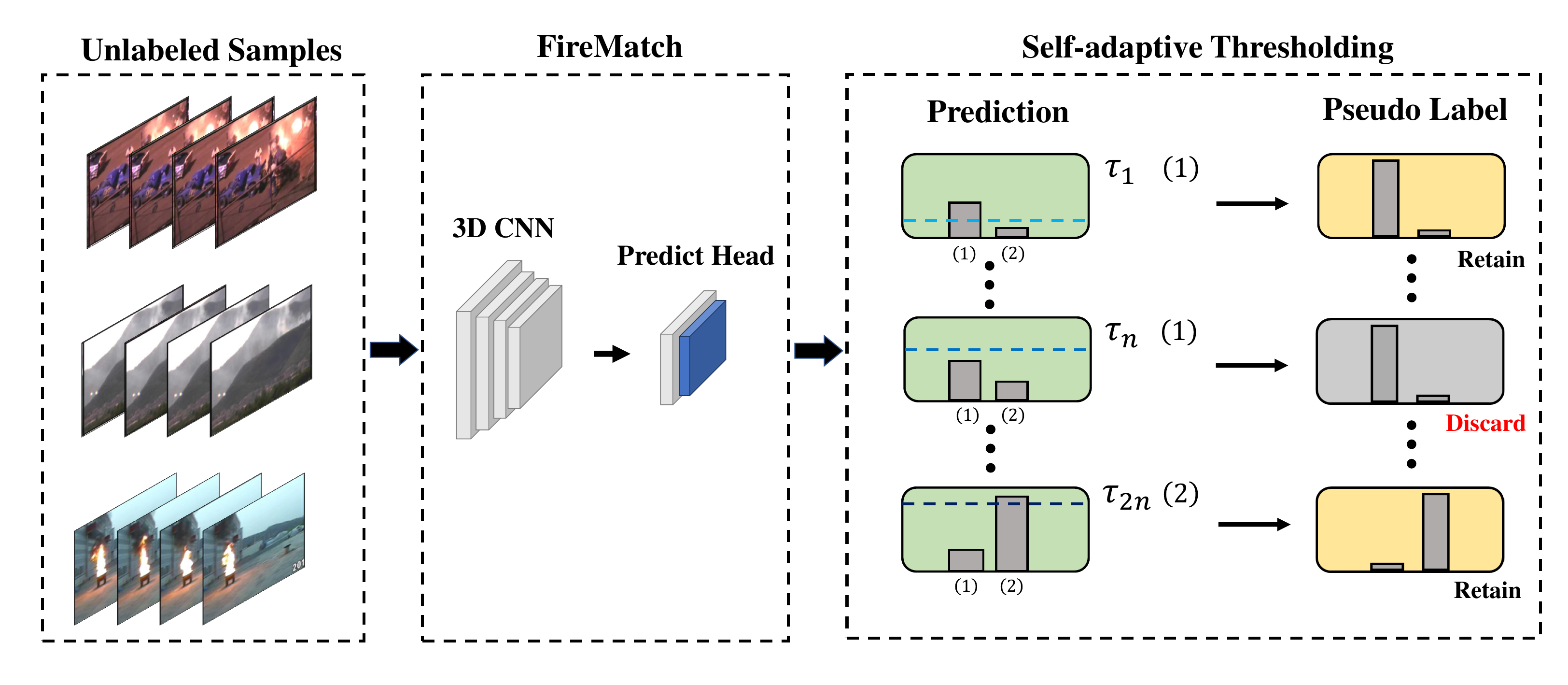}
  \caption{Self-adaptive Thresholding. Unlabeled data is fed into a 3D CNN to extract features and generate predicted results by the Predict Head. The predicted labels with class probabilities higher than the corresponding class threshold are considered as pseudo-label (yellow), while those below the threshold are discarded (gray).}
  \label{SAT}
\end{figure*}

According to the work \cite{WY22}, it is effective to set the pseudo-label threshold based on the learning status of the model. So, we introduce the self-adaptive threshold (SAT) in the part of consistency regularization combined with pseudo-labels. As shown in Figure \ref{SAT}, the SAT adaptively adjusts the confidence threshold of each class during model training according to the learning status of the model, thereby generating trustworthy pseudo-labels.

The motivation behind this method is to set lower thresholds for each class in the early stages of training, which allows more potentially correct samples to be included in training and speeds up convergence. As the model becomes more confident in the later stages of training, the thresholds are raised to filter out incorrect samples. Specifically, the SAT includes both a global threshold and local thresholds. The global threshold is defined as follows:

\begin{equation}
    {\tau _i} = \left\{ {\begin{array}{*{20}{c}}
  {\frac{1}{N},{\text{if }}i{\text{  =  0,}}} \\ 
  {{\lambda _{de}}{\tau _{i - 1}} + \left( {1 - {\lambda _{de}}} \right)\frac{1}{{\mu B}}\sum\limits_{b = 1}^{\mu B} {\max ({q_b}),{\text{  otherwise,}}} } 
\end{array}}\right.
\end{equation}
where $N$ is the number of categories, $i$ represents the number of model iterations, ${\lambda _{de}} \in (0,1)$ represents the EMA momentum decay, $\mu $ remains consistent with Section $A$, and ${q_b}$ is the predicted probability of the model for weakly augmented samples of different categories. The global threshold increases steadily during the training process to ensure the correctness of pseudo-labels.  While the local thresholds are to adjust the global threshold in a class-specific way and defined as:

\begin{equation}
{\tilde p_i}\left( n \right) = \left\{ {\begin{array}{*{20}{c}}
  {\frac{1}{N},{\text{                                                      if }}i = 0,} \\ 
  {{\lambda _{de}}{{\tilde p}_{i - 1}}(n) + (1 - {\lambda _{de}})\frac{1}{{\mu B}}\sum\limits_{b = 1}^{\mu B} {{q_b}(n)} ,{\text{  otherwise,}}} 
\end{array}} \right.
 \end{equation}
where ${\tilde p_i} = \left[ {{{\tilde p}_i}(1),{{\tilde p}_i}(2),...,{{\tilde p}_i}(N)} \right]$ contains all ${\tilde p_i}(n)$. Then, the final threshold can be defined as:

\begin{equation}
    {\tau _i}(n) = MaxNorm({\tilde p_i}\left( n \right)){\text{ }}{\text{}}{\cdot\text{ }\tau _i} 
\end{equation}
where $MaxNorm = \frac{x}{{\max \left( x \right)}}$. By setting adaptive thresholds, we select the class with predicted probabilities higher than the threshold as the pseudo label and further optimize the training process.

\subsubsection{Self-adaptive Fairness}
We further observe that video sample annotation is often much more challenging than image sample annotation.  In practical applications, a shortage of labeled video samples may lead to the model exhibiting higher confidence for certain classes. Specifically, in the binary classification scenario of fire and safety state videos, fire segments usually come after safety state segments, leading the model to have higher confidence in the non-fire class which affects the model's classification ability. Therefore, we introduce a fair class objective to encourage the model to make different predictions. We normalize the histogram distribution of the pseudo labels based on the expectation of probabilities:
\begin{equation}
\begin{gathered}
  \bar p = \frac{1}{{\mu B}}\sum\limits_{b = 1}^{\mu B} {\mathbbm{1}(\max ({q_b}) \geq {\tau _i}(\arg \max ({q_b}))} {Q_b}, \hfill \\
  \bar h = His{t_{\mu B}}\left( { \mathbbm{1} (\max ({q_b}) \geq {\tau _i}(\arg \max ({q_b})){{\hat Q}_b}} \right), \hfill \\
  {{\tilde h}_i} = {\lambda _m}{{\tilde h}_{i - 1}} + (1 - {\lambda _m})His{t_{\mu B}}({{\hat q}_b}). \hfill \\ 
\end{gathered} 
\end{equation}
where ${q_b}$ and ${Q_b}$ denote the predicted probability of the model for a weak and strong augmented sample, ${\hat q_b}$ and ${\hat Q_b}$ represent the corresponding ``one-hot" labels. The self-adaptive fairness loss can be defined as:

\begin{equation}
    {\ell _{fair}} =  - {\mathcal{H}}(SumNorm(\frac{{{{\tilde p}_i}}}{{{{\tilde h}_i}}}),SumNorm(\frac{{\bar p}}{{\bar h}})).
\end{equation}
where $SumNorm = ( \cdot )/\sum {( \cdot )}$. 
We employ this fairness objective to optimize the model, enabling it to mitigate biases towards specific classes and promote more precise predictions.

\subsection{Adversarial Distribution Alignment}
\subsubsection{Video Cross-set Sample Augmentation}
\begin{figure*}[!t]
  \centering
  \includegraphics[width=6in]{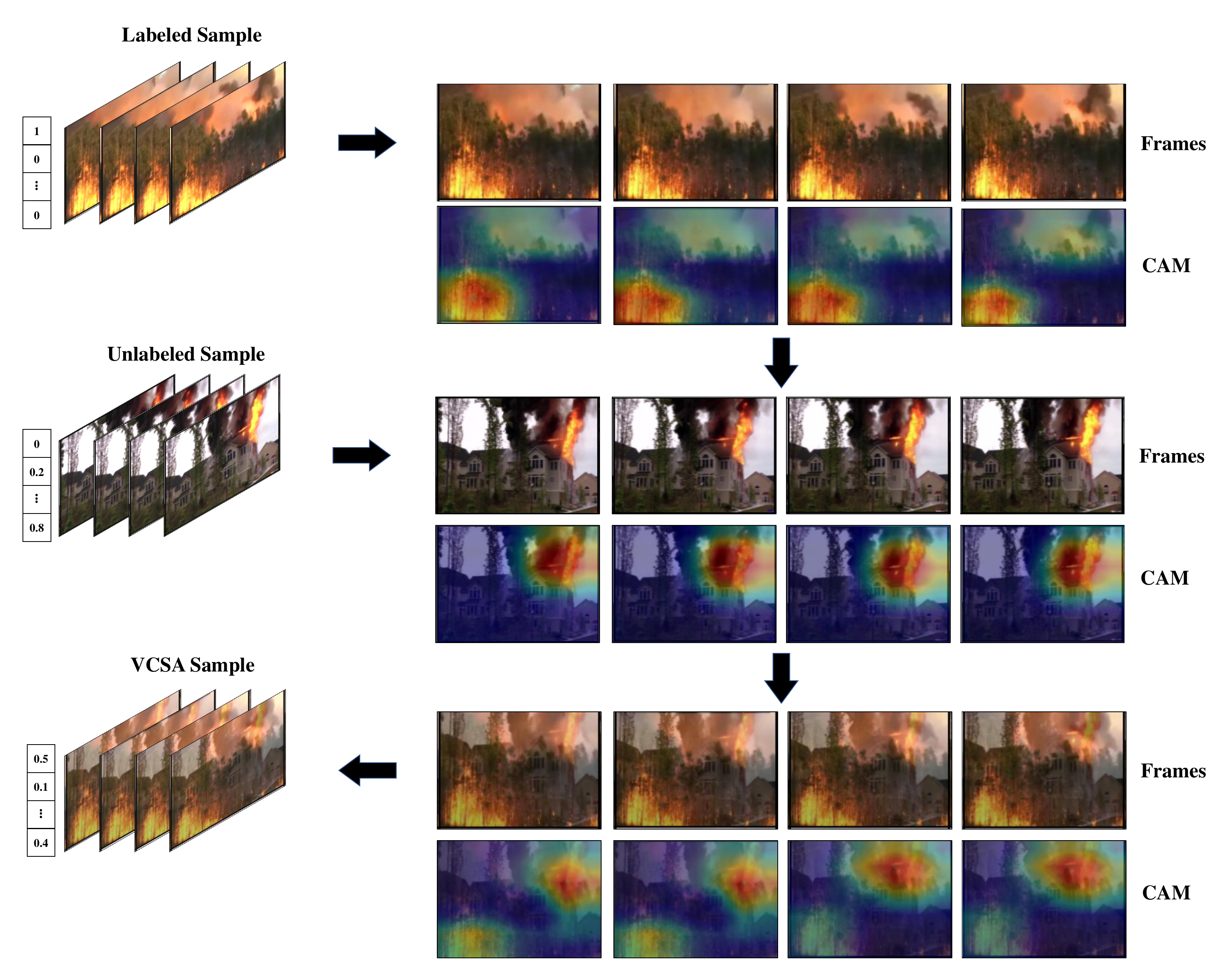}
  \caption{Class activation mapping (CAM) \cite{ZB16} on samples. We briefly show the generation of Video Cross-set Sample Augmentation (VCSA) sample and the spatio-temporal CAM score map of each sample.}
  \label{TCSA}
\end{figure*}

In semi-supervised learning, the limited sampling of labeled data affects model optimization greatly and leads to decreased classification performance \cite{WQ19}. We recommend fully utilizing both labeled and unlabeled samples by generating new effective training samples through interpolation, as shown in Figure \ref{TCSA}. 
To achieve interpolate augmentation, we reshape the ${x_b}$ and ${u_b}$ into $x_b^m{\text{ }} \in {\mathbb{R}^{K \times W \times H}}$ and $u_b^m \in {\mathbb{R}^{K \times W \times H}}$ respectively, where $K = T \cdot C$. Then, the interpolated samples can be represented as:
\begin{equation}
\label{Cross}
    \begin{gathered}
  {{\tilde x}_b} = {\lambda _m}\cdot x_b^m + (1 - {\lambda _m})\cdot u_b^m{\text{ }}, \hfill \\
  {{\tilde y}_b} = {\lambda _m}\cdot y_b^{} + (1 - {\lambda _m})\cdot{{y'}_b}{\text{ }}, \hfill \\
  {{\tilde z}_b} = {\lambda _m}\cdot 0 + (1 - {\lambda _m})\cdot 1{\text{ }}. \hfill \\ 
\end{gathered} 
\end{equation}   
where ${\lambda _m}$ is a random variable generated based on the prior $\beta$ distribution of $\beta (\alpha ,\alpha )$, and the hyperparameter $\alpha $ controls the shape of the distribution of $\beta $. Additionally, ${\tilde x_b}$ and ${\tilde y_b}$ represent the interpolated augmentation sample and its corresponding class label respectively, and the corresponding label of discriminator ${\mathcal{D}}$ is ${\tilde z_b}$.

We refer to this type of cross-set augmentation as Video Cross-set Sample Augmentation (VCSA). The motivation is to take into account the temporal information present in video data, as the formation of fire in fire scenes always starts small and grows bigger, and even in intense explosion scenes, there is a gradual increase in the number of flame pixels in a few frames. Therefore, preserving the temporal relationships between frames in cross-set interpolation augmentation is beneficial for generating more informative samples. In fact, the proposed video augmentation strategy is strongly correlated with the \textit{Mixup} \cite{ZH17} method. It can be regarded that for each frame of both labeled and unlabeled videos, we apply the \textit{Mixup} method to generate new augmented data, and extend it to the domain of semi-supervised video classification. In addition, the work \cite{ZH17} shows that cross-set sample augmentation greatly expands the set of valuable training data, making the learning process more stable and improving the robustness of the model.  With regard to the distribution of samples, the work \cite{WQ19} proves that the distribution of pseudo-samples is closer to the true distribution than that of original labeled samples.

\subsubsection{Distribution Distance Minimization}
In Section $3.3.1$, we have described in detail the data augmentation method VCSA applied to adversarial distribution alignment. The data generated from VCSA will be fed back to the 3D CNN for feature extraction. Based on the features extracted by the 3D CNN, we optimize the model using the adversarial distribution alignment strategy. Intuitively, when the distribution gap between labeled and unlabeled data is large, the discriminator can easily distinguish between the two, resulting in a small prediction error, and vice versa. To reduce the distribution mismatch between labeled and unlabeled data, we minimize the distribution distance, forcing the 3D CNN feature extractor to generate features that are well-aligned in the latent space. Additionally, we also set up a classifier to generate pseudo-labels for the next iteration of interpolation fusion. Therefore, the training objective of the adversarial distribution alignment stage is as follows:
\begin{equation}
\begin{split}
    {\mathcal{L}_{align}} = \mathop {min}\limits_{f,cls,\mathcal{D}} \sum\limits_{\tilde x} {\rho {\text{ }} \cdot } {\text{ }}\mathcal{H}\left( {cls\left( {f\left( {{{\tilde x}_b}} \right)} \right),{{\tilde y}_b}} \right) + \\ {\lambda _m} \cdot \mathcal{H}\left( {\mathcal{D}\left( {f\left( {{{\tilde x}_b}} \right)} \right),{{\tilde z}_b}} \right).
\end{split}
\end{equation}
where $f( \cdot )$ , $cls( \cdot )$ , and $\mathcal{D}( \cdot )$ refer to the feature extractor, classifier, and discriminator, respectively. $\mathcal{H}( \cdot {\text{ }},{\text{ }} \cdot )$ represents the cross-entropy loss function. $\rho$ and ${\lambda _m}$ are the weight of the classification loss, where ${\lambda _m}$ consistent with Eq. \ref{Cross}. Generally, when   
${\lambda _m}$ is larger, the proportion of labeled data $\tilde x$ is higher, and the corresponding label ${\tilde y}$ have higher credibility.

\subsection{Loss Function}
The final loss of the proposed method consists of five components: the classifier loss ${{\mathcal{L}}_{cs}}$ and the prediction head loss ${{\mathcal{L}}_{ps}}$ for labeled data, the adversarial distribution alignment loss ${{\mathcal{L}}_{align}}$ and consistency loss ${{\mathcal{L}}_{match}}$ for unlabeled data, the fairness loss ${{\mathcal{L}}_{fair}}$ to drive model to make balanced predictions for each class. We define ${p_{cs}}\left( {y|{x_b}} \right)$ and ${p_{ps}}\left( {y|{x_b}} \right)$ as the probabilities predicted by the classifier and the prediction head for input ${x_b}$. Therefore, the definitions of the supervised loss are as follows:
\begin{equation}
    \begin{gathered}
  {{\mathcal{L}}_{cs}} = \frac{1}{B}\sum\limits_{b = 1}^B {{\mathcal{H}}({y_b},{p_{cs}}(y|{x_b}))} {\text{ }}, \hfill \\
  {{\mathcal{L}}_{ps}} = \frac{1}{B}\sum\limits_{b = 1}^B {{\mathcal{H}}({y_b},{p_{ps}}(y|{x_b}))} {\text{ }}. \hfill \\ 
\end{gathered} 
\end{equation}
Then, the adversarial distribution alignment loss is defined as:
\begin{equation}
\begin{split}
    \mathcal{L}_{\text{align}} = \frac{1}{B}\sum_{b = 1}^B \rho \cdot \mathcal{H}\left(\tilde{y}_b, p_{cs}(y|\tilde{x}_b)\right) + \\
    \lambda_m \cdot \mathcal{H}\left(\tilde{z}_b, p_d(z|\tilde{x}_b)\right).
\end{split} 
\end{equation}
where $\tilde{x}_b$ is VCSA sample, $\tilde{y}_b$ are corresponding labels, $\tilde{z}_b$ denotes discriminator labels, and $p_d(z|\tilde{x}_b)$ represents discriminator output result for $\tilde{x}_b$. Then, the consistency loss can be defined as:
\begin{equation}
\begin{split}
    \mathcal{L}_{\text{match}} = \frac{1}{\mu B}\sum_{b = 1}^{\mu B} \left(\mathbbm{1}(\max(q_b) > \tau_i(\arg \max (q_b))) \cdot \mathcal{H}(\hat{q}_b, Q_b)\right).
\end{split}
\end{equation}
where ${Q_b} = {p_p}\left( {y|s{u_b}} \right)$ and ${q_b} = {p_p}\left( {y|w{u_b}} \right)$ are predict result of predict head to strong and weak augmentation samples, respectively. ${\hat q_b}$ denote “one-hot” label for ${q_b}$. So, the final loss of proposed method is defined as:
\begin{equation}
\begin{split}
    {\mathcal{L}} = {\omega _m}({{\mathcal{L}}_{ps}} + {{\mathcal{L}}_{match}}) + {\omega _f}{{\mathcal{L}}_{fair}} + \\ {\omega _a}{{\mathcal{L}}_{align}} + {{\mathcal{L}}_{cs}}.
\end{split}
\end{equation}
where ${\omega _m}$, ${\omega _f}$, and ${\omega _a}$ are the loss weight for the consistency loss, fairness loss, and adversarial distribution alignment loss, respectively.

\section{EXPERIMENT}
This chapter provide a detailed description of the experimental details. Section 4.1 presents the specific settings and hardware devices used in the experiments. Section 4.2 mainly introduces the datasets used in the experiments. Section 4.3 provides an analysis of the comparison results between the proposed algorithm and the state-of-the-art semi-supervised classification algorithms. Section 4.4 shows the results of the ablation experiments.

\subsection{Implement Details}
The initial learning rate $\eta$ is set to 0.03 and the cosine learning rate decay \cite{IL16} is used to update the learning rate to $\eta \cos \left( {\frac{{7\pi k}}{{16K}}} \right)$, where $k$ is the current iteration number and $K$ is the total number of training steps. In previous works \cite{BD20,SK20,ZB21}, it has been shown that the Adam optimizer \cite{KDP14} results in worse performance in similar tasks, therefore we use SGD with momentum \cite{SI13} instead. For fairness in the experiments, we uniformly set the backbone of all algorithms to the 3D-ResNet \cite{KH20}, and use the same weight decay of 0.0005, the batch size $bs = 6$ for labeled data and unlabeled data batch size ratio $\mu  = 4$ . The network is  trained for 500 epochs with a fixed random seed to obtain the optimal result. Additionally, we implement all methods in Python 3.9 and PyTorch 1.12.1, and MegEngine 1.12 \footnote{https://github.com/MegEngine/MegEngine} is used to save memory space. A computer equipped with Ryzen 7 5800X 3.8 GHz (32GB RAM) and NVIDIA GeForce RTX 3090 is used for training and testing.

\begin{figure*}
  \centering
  \includegraphics[width=\textwidth]{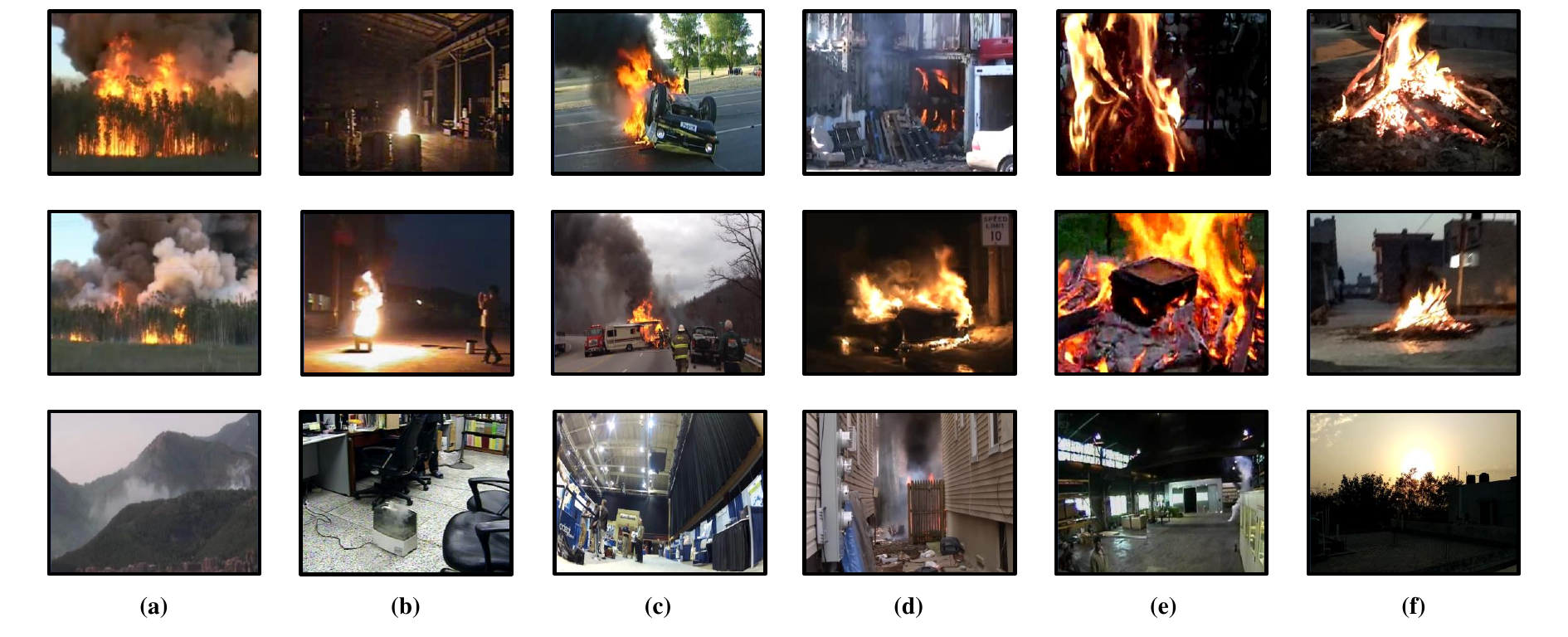}
  \caption{Examples of fire-related datasets. (a) MIVIA Fire and Smoke. (b) KMU Fire and Smoke Database. (c) Furg Fire Dataset. (d) LAD2000. (e) Firesense. (f) Custom-Compiled Fire Dataset.}
  \label{Datasets}
\end{figure*}

\subsection{Datasets}
Since the publicly available datasets for fire video classification currently contain limited video data, we integrate existing fire-related datasets to facilitate training and testing. The datasets are as follows:

\textbf{MIVIA Fire and Smoke} \cite{DLR14,FP15}: We combine the MIVIA fire detection dataset and smoke detection dataset as a new dataset, which contains 180 videos captured from the real world. The dataset mainly includes 14 videos with flame and 166 videos without any events of interest. Specifically, among these 166 non-fire videos, there are key situations traditionally associated with fires, such as moving red objects, smoke, or clouds in the scene.

\textbf{KMU Fire and Smoke Database} \cite{KBC11}: This dataset contains 22 short-distance outdoor flame videos, 2 indoor short-distance smoke videos, 4 wildfire smoke videos, and 10 videos of moving objects resembling smoke or flames.

\textbf{Furg Fire Dataset} \cite{HV17}: The dataset contains 17 videos with fire, including footage of burning cars and house fires. Additionally, the dataset includes 6 non-fire videos unrelated to fire incidents.

\textbf{LAD2000} \cite{WB21}: This dataset is a large-scale video dataset for anomaly detection, consisting of 2000 video sequences across 14 categories such as fire, fighting, and destruction in real-world settings. We select 107 videos with fire, which contain a variety of fire situations and different types of wildfires and human-caused fires in diverse scenes.

\textbf{Firesense} \cite{KD15}: This dataset is commonly used for automatic flame detection in videos. We collect a total of 39 videos from this dataset, including 16 fire videos and 23 non-fire videos related to fire incidents, such as strong light sources, moving smoke, and bright outdoor environments.

\textbf{Custom-Compiled Fire Dataset} \cite{JA19}: This is a real-world fire detection dataset, with a designated test set that includes 46 fire videos and 16 non-fire videos. These data are diverse enough to match most fire situations well.

To mitigate the impact of imbalanced data on model classification performance and further evaluate the model's generalization ability, we integrate MIVIA Fire and Smoke \cite{DLR14,FP15}, KMU Fire, and Smoke Database \cite{KBC11}, Furg Fire Dataset\cite{HV17}, and LAD2000 \cite{WB21} as the training set, including a total of 164 video sequences with flame and 184 non-fire video sequences. Firesense \cite{KD15} and Custom-Compiled Fire Dataset \cite{JA19} are both serve as validation and test sets. We unify all video data to a resolution of 320×240 and a frame rate of 30 fps to ensure consistency. We summarize the datasets involved in this paper in Table \ref{Data_Table}.

\begin{table}
\centering
\caption{Detailed Configuration of the Dataset.}
\begin{tabular}{c|c|c|c}
\toprule
Datasets                      & Fire    & Non-Fire    & Total \\
\midrule
MIVIA Fire and Smoke          & 14      & 166         & 180    \\
KMU Fire and Smoke Database   & 26      & 12          & 38 \\
Furg Fire Dataset             & 17      & 6           & 23    \\
LAD2000                       & 107     & -           & 107    \\
\midrule
\textbf{Training set}         & 164     & 184         & 348     \\
\midrule
Firesense                     & 23      & 16          & 39    \\
Custom-Compiled Fire Dataset  & 45      & 16          & 61   \\
\bottomrule
\end{tabular}
\label{Data_Table}
\end{table}

The cumulative frames in the training set have reached 51,174, comprising 23,442 fire video frames and 27,732 non-fire video frames. Additionally, the Custom-Compiled Fire Dataset includes 4,335 fire video frames and 1,608 non-fire video frames, while the Firesense dataset contains 2,215 fire video frames and 2,881 non-fire video frames.

\begin{table*}[!t]
\centering
\caption{Accuracy (\%) of semi-supervised classification methods on Firesense and Custom-Compiled Fire Dataset. The best results show in \textcolor{red}{\underline{\textbf{red}}} and second-best result show in \textcolor{blue}{blue}.}
\resizebox{\textwidth}{!}{
\begin{tabular}{clccclccc}
\toprule
Dataset   &  & \multicolumn{3}{c}{Firesense}    &  & \multicolumn{3}{c}{Custom-Compiled Fire Dataset} \\ \cmidrule{1-1} \cmidrule{3-5} \cmidrule{7-9} 
Method    &  & 72 labels (20\%) & 36 labels (10\%) & 18 labels (5\%) &  & 72 labels (20\%)       & 36 labels (10\%)      & 18 labels (5\%)      \\ 
\midrule
Mean Tearcher  &  & 66.67 & 64.10 & 61.54  &  & 80.33 & 73.77 & 72.13 \\
ADA-Net     &  & \textcolor{blue}{74.36}     & 66.67     & 64.10    &  & 86.89           & 85.25          & 81.97         \\
FixMatch   &  & 71.79     & 69.23     & 61.54    &  & 80.33           & 77.05          & 75.41         \\
FreeMatch  &  & 69.23     & 64.10     & 61.54    &  & \textcolor{blue}{90.16}           & 85.25          & 78.69         \\
FlexMatch  &  & \textcolor{blue}{74.36}     & \textcolor{blue}{71.79}     & \textcolor{blue}{69.23}    &  & 88.52           & \textcolor{blue}{86.89}          & \textcolor{blue}{83.61}         \\
UPS        &  & 66.67     & 64.10     & 61.54    &  & 88.52           & 80.33          & 77.05         \\ 
\textbf{FireMatch}       &  & \textcolor{red}{\underline{\textbf{76.92}}}     & \textcolor{red}{\underline{\textbf{74.36}}}     & \textcolor{red}{\underline{\textbf{71.79}}}    &  & \textcolor{red}{\underline{\textbf{91.81}}}           & \textcolor{red}{\underline{\textbf{88.52}}}          & \textcolor{red}{\underline{\textbf{83.61}}}         \\
\bottomrule
\end{tabular}}
\label{Comparison}
\end{table*}

\subsection{Baseline Methods}
We remold existing state-of-the-art classification methods based on semi-supervised learning to fully match the fire video classification task.

\textbf{Mean Teacher} \cite{TA17} enhances neural network training by introducing a ``teacher" network that enforces consistency with a ``student" network. It encourages similar predictions on both labeled and unlabeled data, improving generalization in semi-supervised learning.

\textbf{FixMatch} \cite{SK20} combines consistency regularization with pseudo-labeling to improve the classification performance of a portion of the model using a small amount of labeled data first. For unlabeled data, FixMatch performs strong and weak augmentations separately. Weakly augmented samples of unlabeled data are used for predicting pseudo-labels, which are retained when the predicted probability is higher than 0.95, and discarded otherwise. Based on the assumption of consistency regularization, the strongly augmented samples input to the model are encouraged to predict the pseudo-labels of the weakly augmented samples, to learn a more robust feature representation.

\textbf{FreeMatch} \cite{WY22} is based on the same core idea as FixMatch but with a different approach involving a dynamic threshold adjustment strategy. In the early stages of training, a lower pseudo-label threshold is set to obtain more pseudo-labeled samples and accelerate convergence. As the improvement of model's classification ability, the threshold gradually increases to filter out incorrect pseudo-labels and improve classification accuracy.

\textbf{FlexMatch}  \cite{ZB21} believes that the threshold for pseudo-labeling should be adjusted based on the model's learning state for each sample. When the classification accuracy of a certain class is low, the model is not satisfied with the learning state of that class. Therefore, a low threshold encourages the model to learn more samples from that class. When the threshold for pseudo-labeling is high, the learning effect of a certain class can be judged by the number of samples that fall into that class and exceed the threshold. Based on the above ideas, FlexMatch proposes a curriculum pseudo-label to flexibly adjust the thresholds for different categories and select informative unlabeled data to improve the model's classification performance.

\textbf{ADA-Net} \cite{WF22} is an enhanced distribution alignment network, which effectively limits the generalization error of semi-supervised learning by minimizing the training error of labeled data and the empirical distribution gap between labeled and unlabeled data. To ensure a fair comparison, the 3D version of ADA-Net adopt in our experiments uses the same video data augmentation method as FireMatch.

\textbf{UPS} \cite{RMN21} is a typical semi-supervised classification algorithm based on pseudo-labeling. Its uncertainty-based pseudo-label selection framework significantly reduces the noise encountered during the training process to improve the accuracy of pseudo-labels. In addition, UPS allows for the creation of negative pseudo-labels to improve negative learning in single-label classification.

\subsection{Experimental Results and Analysis}
We conduct extensive experiments on Firesense \cite{KD15} and Custom-Compiled Fire Dataset \cite{JA19}. Three labeled data amounts  (72 labels, 36 labels, and 18 labels) are set to evaluate the different semi-supervised classification methods using top-1 accuracy as the evaluation metric. 

The experimental results are shown in Table \ref{Comparison}. We regard the Mean Teacher as the baseline model for our experiments. From the experimental results, it appears that its performance is not satisfactory under several different label quantity settings. The reason for this phenomenon might be that the Mean Teacher method has a relatively high requirement for the quality of labeled data, whereas these labeled fire video data used in training can be considered weakly supervised to some extent. Our method achieves the best classification results on the Firesense dataset, with accuracies of 76.92\% (72 labels), 74.36\% (36 labels), and 71.79\% (18 labels), respectively. The results indicate that reducing the number of labeled data does not have a significant impact on FireMatch, which can be attributed to the correct prediction of unlabeled data and the involvement of diversified augmented samples in the training process. In addition, ADA-Net and FlexMatch both achieve suboptimal results of 74.36\% with 72 labels. However, once the number of labeled data is reduced, ADA-Net's performance on this dataset is severely affected, while FlexMatch still maintains its suboptimal results. This is because ADA-Net needs correct labeled samples to align the distribution of labeled and unlabeled data, and once the number of labeled samples is reduced, it is difficult to reduce the distribution gap between the two. Comparing with the ADA-Net, FlexMatch can generate more correct pseudo-labels for unlabeled data based on its unique CPL strategy. FreeMatch and FixMatch rely too much on labeled samples to enhance the model's classification ability, and then accept pseudo-labels for unlabeled samples at different thresholds. Therefore, when the number of labeled samples drops to 18, both methods only achieve a classification accuracy of 61.54\%. Moreover, UPS do not achieve the expected results on Firesense, with classification accuracies of 66.67\%, 64.10\%, and 61.54\% under the three label quantity settings, respectively. This is probably because the Firesense dataset contains too many confusing samples (e.g., strong light sources, moving smoke), and the distribution of these samples differs significantly from that of the training data, which limits UPS performance.

\begin{figure}[!t]
\centering
\includegraphics[width=3.4in]{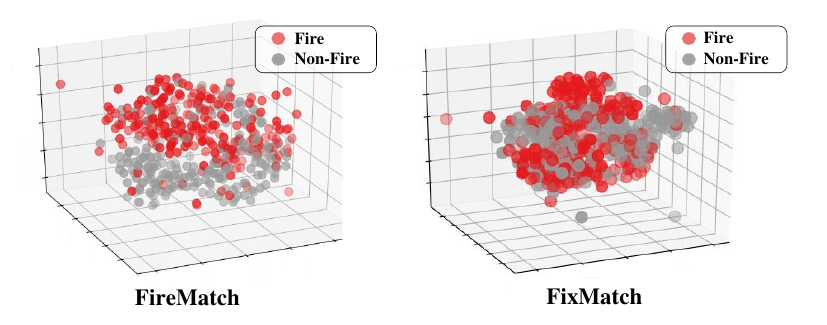}
\caption{The t-SNE (Perplexity:30, LR:200, Iter:1000) visualization of the features extracted from the last conv layer of FireMatch and FixMatch.}
\label{TSNE}
\end{figure}

Unlike Firesense, Custom-Compiled Fire Dataset has a sufficiently diverse set of samples, allowing various baseline methods to achieve decent classification results. FireMatch still achieves 91.81\%, 88.52\%, and 83.61\% accuracy under the three label settings, respectively. However, UPS and FlexMatch both achieved 88.52\% classification accuracy under the 72 labels setting. We also provide visualizations of the feature extraction results from the last convolutional layer of FireMatch, as shown in Figure \ref{TSNE}. It can be seen that FireMatch can effectively distinguish between fire and non-fire video samples, whereas FixMatch does not exhibit high discriminability between the two classes in the distribution space. UPS improves its classification performance by continuously iterating and selecting correct pseudo-labeled samples during training. However, when the labeled sample count decreases to 36 or 18, UPS's classification ability decreases significantly, resulting in a substantial drop in classification accuracy due to the inability to select more correct pseudo-labels during iteration. ADA-Net achieves a classification accuracy of 86.89\% under the 72 labels setting, which is 4.92\% and 1.63\% lower than FireMatch and FlexMatch, respectively.  However, when the label count is 36, ADA-Net still maintains a decent classification accuracy of 85.25\%. This may be because the cross-set augmented sample distribution generated in ADA-Net is closer to the distribution of the test set data, enabling correct classification of test samples. The results achieved by FixMatch and FreeMatch are not satisfactory, as their threshold adjustment strategy cannot fully leverage its classification ability when the training sample count is low, especially when the labeled sample count is low.

\begin{figure*}
    \centering
    \includegraphics[width=0.8\textwidth]{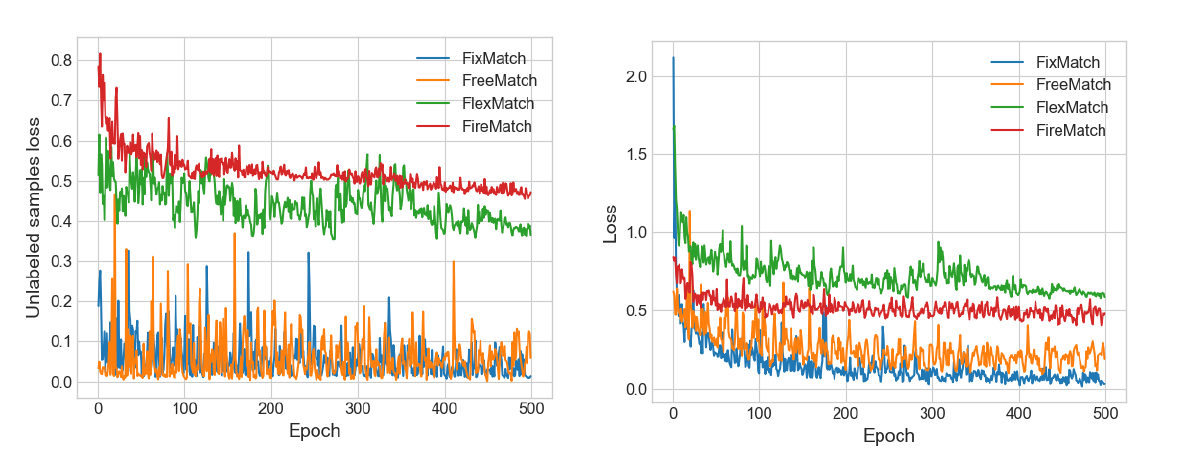}
    \caption{Unlabeled data loss (left) and overall loss (right) of the Match family of semi-supervised classification methods. Compared to other methods based on consistency regularization, FireMatch exhibits a relatively stable training process with unlabeled data, thereby reducing fluctuations throughout the overall training process.}
\label{Loss}
\end{figure*}

Semi-supervised classification methods based on consistency regularization are susceptible to the instability caused by random augmentation in their unlabeled data. In Figure \ref{Loss}, we present the unlabeled data loss and overall loss of several Match family of algorithms during the training process. FireMatch exhibits a more stable training of unlabeled data, followed by FlexMatch with some fluctuations. FixMatch and FreeMatch show significant fluctuations in the training of the unlabeled data. This directly affects the stability of semi-supervised classification algorithms based on consistency regularization during training, FireMatch demonstrates a more stable loss curve and faster convergence speed. Similarly, FlexMatch is less affected by the unlabeled data, while FixMatch and FreeMatch are affected greatly, resulting in slower convergence.

\begin{table}
\centering
\caption{The ablation study of FireMatch.}
\begin{tabular}{c|l|c}
\toprule
Index      & Method                         & Accuracy (\%) \\
\hline
1          &   CR+FT                        & 80.33    \\
2          &   CR+SAT                       & 85.25 \\
3          &   CR+SAT+ADA(VM)               & 86.89    \\
4          &   CR+SAT+ADA(VCSA)             & 88.52    \\
5          &   CR+SAT+ADA(VCSA)+FL          & \textcolor{red}{91.81}    \\
\hline
6          &   FireMatch (RCF)              & 90.16    \\
7          &   FireMatch (OF)               & \textcolor{red}{91.81}    \\
8          &   FireMatch (Sharpen)          & 88.52  \\
9          &   FireMatch (Smooth)           & 83.61  \\
\hline
10          &   Fully-Supervised (ResNet18)            & 90.16    \\
11          &   Supervised (ResNet18)     & 86.89  \\
\bottomrule
\end{tabular}
\label{Ablation}
\end{table}

\subsection{Ablation Study}
 In this section, we conduct a series of ablation experiments on the Custom-Compiled Fire Dataset with 20\% labeled data to validate the effectiveness of FireMatch components. Table \ref{Ablation} shows the results of different component ablation experiments, where ``CR+FT" represents the pseudo-labeling method using only consistency regularization (CR) and fixed threshold (FT). ``CR+SAT" represents the pseudo-labeling strategy combining CR with self-adaptive threshold (SAT). ``CR+SAT+ADA(VM)" represents the data branch with adversarial distribution alignment (ADA) added on the basis of CR and SAT, where the video enhancement strategy used by ADA is VideoMix (VM) \cite{YS20}. Additionally, ``VCSA" represents the video cross-set sample augmentation proposed in this paper, and ``+FL" indicates the addition of the fairness loss (FL).

\begin{figure}[thbp]
\centering
\includegraphics[width=3.2in]{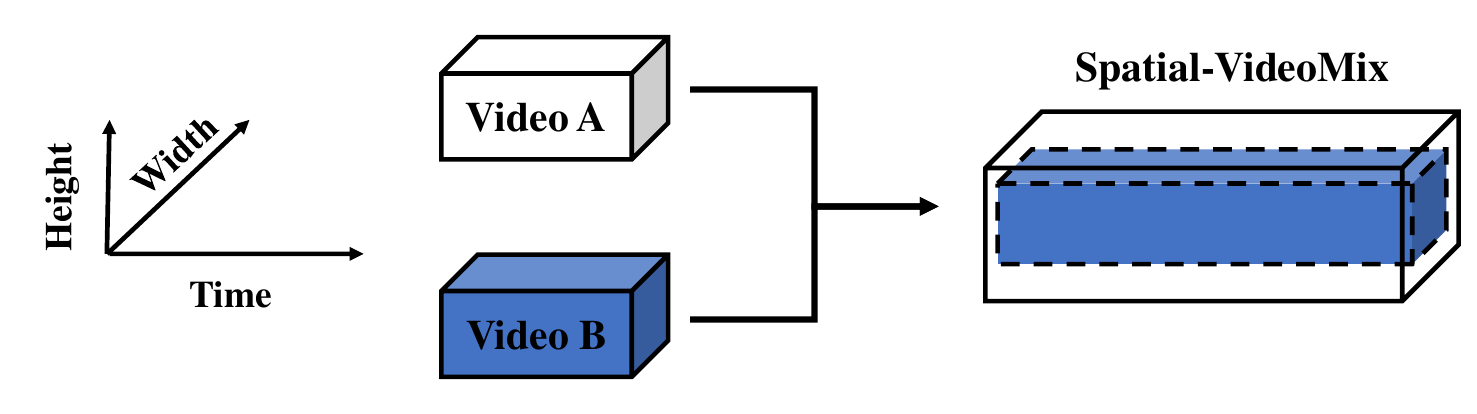}
\caption{Spatial-VideoMix used in ablation study. The sub-cube of video B (labeled video data represented by the blue cube) is inserted into video A (unlabeled video data represented by the white cube) to generate new video-augmented samples.}
\label{VM}
\end{figure}

The experimental results show that SAT significantly improves classification accuracy. 
Furthermore, based on CR and SAT, we utilize ADA as a means of augmenting training samples to reduce the discrepancy in sampling empirical distributions and enhance classification accuracy. Additionally, we provide ablation results of the ADA strategy based on VM. As shown in Figure \ref{VM}, VM is a useful video data augmentation method, but in this work, the enhancement mode of VM can cause occlusion of the target flame, resulting in suboptimal classification results.  FL further improves the classification ability of the model by penalizing monotonous prediction results at the training stage.

We also conduct ablation experiments on the hypothesis proposed in Section \MakeUppercase{\romannumeral3} and compare our method with supervised 3D-ResNet. ``FireMatch (RCF)" represents weak augmentation involving random cropping and flipping of video frames in consistency regularization. ``FireMatch (OF)" represents flipping only for weakly augmented video data. ``FireMatch(Sharpen)" and ``FireMatch(Smooth)" respectively denote weak augmentation modes using sharpening and smoothing. These two forms of weak augmentation do not yield favorable results. We believe that sharpening increases the strength of noise in the data while smoothing results in the loss of fire-related features, which is disadvantageous for fire detection in videos.
``Fully-Supervised (ResNet18)" represents training a fully supervised 3D-ResNet classification model as backbone with all data labels provided. Similarly, ``Supervised (ResNet18)" denotes using only 72 labeled data. The experimental results validate our hypothesis proposed in Section \MakeUppercase{\romannumeral3}, and it can be seen that our proposed method outperforms the supervised 3D-ResNet in the 20\% label setting. This is reasonable as our training and testing data come from different sources, and there are significant differences in data distribution. 3D-ResNet can learn the feature representation of the training data well, but its classification performance is affected when facing test data with significant differences. Besides, FireMatch can produce correct pseudo-labels for unlabeled data and improve the model's generalization performance by synthesizing some augmentation samples with the correct label.

\begin{table}
\centering
\caption{The evaluation of different 3DCNNs as backbone for feature extraction.}
\begin{tabular}{c|l|c|c}
\toprule
Index      & Backbone            & Parameters (M)           & Accuracy (\%) \\
\hline
1          &   MobileNetV2       & 4.73                    & 77.05    \\
2          &   EfficientNet       & 7.06                       & 78.69    \\
3          &   ShuffleNetV2       & 3.41                    & 85.25    \\
4          &   SqueezeNet         & 3.95                   & 80.33    \\
\hline
5          &   DenseNet          & 13.40                       & 88.52    \\
6          &   ResNeXt18         & 18.13                   & 90.16 \\
7          &   ResNet10          & 16.51                   & 90.16  \\
8          &   ResNet18          & 34.79                   & \textcolor{red}{91.81}  \\
\bottomrule
\end{tabular}
\label{Backbone}
\end{table}

Additionally, we study the impact of different backbone networks on the algorithm's performance. 3D versions of MobileNetV2 \cite{KOR19}, EfficientNet \cite{TME19}, ShuffleNetV2 \cite{KOR19}, SqueezeNet \cite{KOR19}, DenseNet \cite{HGD17}, ResNeXt \cite{XSA17}, and ResNet \cite{KOR19} serve as backbone networks for feature extraction. Under consistent configurations, we retrain them to obtain various models. The experimental results are presented in Table \ref{Backbone}. We observed that lightweight network models such as MobileNetV2, ShuffleNetV2, and SqueezeNet constrain the performance of FireMatch. In contrast, when using ResNeXt18 and ResNet10 as backbone networks, FireMatch achieves a classification accuracy of 90.16\%. Notably, ResNet18 exhibits the highest classification accuracy at 91.81\%. From the above experiments, we can draw a conclusion that model's parameter size and network depth bring certain advantages to video fire classification.

\section{Limitation and Future Work}
Although FireMatch has shown impressive performance on the current fire detection datasets, it is still far from perfect. We would like to point out some shortcomings in our work to help other researchers advance the field. Firstly, FireMatch is a video-based fire classification framework, which incurs significantly higher computational costs compared to image-based classification models. Training a more robust FireMatch necessitates access to greater computing resources. Secondly, the scale of available datasets for fire video detection is currently limited, which somewhat diminishes the persuasiveness of our proposed approach. Lastly, fire incidents represent only one category of numerous safety events, and there is a lack of reliable and comprehensive safety event datasets that could facilitate the development of video-based classification models for various safety incidents, such as blocked fire exits, leaking roofs, and collisions. Given these considerations, we have already collaborated with relevant enterprises to collect a portion of authentic safety event video data. 
In the future, we will explore the application of knowledge distillation to semi-supervised video classification for the purpose of lightweighting network models and apply this approach to various video classification tasks. Simultaneously, we will construct a high-quality dataset for safety incident classification in hub-level logistics scenarios, further expanding the practical applications of video-based semi-supervised safety event detection.

\section{Conclusion}
In this paper, we propose FireMatch, a semi-supervised fire detection model. In semi-supervised classification tasks, the most important thing is how to make full use of limited labeled data and a large amount of unlabeled data.  To fully utilize the unlabeled data, we first combine consistency regularization with pseudo-labeling. Secondly, we use adversarial distribution alignment to leverage labeled and unlabeled data and generate video cross-set augmentation samples closer to the real distribution to improve the model's generalization performance for different distribution data. Finally, to address the problem of confidence bias in fire video classification, we introduce a fairness loss and encourage the model to make diverse predictions during training. In summary, FireMatch achieves accurate video fire classification and provides a innovative idea for future research.

\section*{Acknowledgements}
This work is partially supported by the National Key Research and Development Program of China (Grant No. 2022YFC3302200) and National Natural Science Foundation of China (61972187, 62276146).


\end{document}